\documentclass{article}

\usepackage{arxiv}

\usepackage[utf8]{inputenc} 
\usepackage[T1]{fontenc}    
\usepackage{hyperref}       
\usepackage{url}            
\usepackage{booktabs}       
\usepackage{amsfonts}       
\usepackage{nicefrac}       
\usepackage{microtype}      
\usepackage{lipsum}
\usepackage{graphicx}
\graphicspath{ {./images/} }

\usepackage{graphicx}
\usepackage[english]{babel}     

\title{Lightweight Structured Multimodal Reasoning for Clinical Scene Understanding in Robotics}

\author{
 Saurav Jha\\
  SETLabs Resarch GmbH\\
  80686 Munich, Germany \\
   \And
    Stefan K. Ehrlich\\
  SETLabs Resarch GmbH\\
  80686 Munich, Germany \\
  \texttt{stefan.ehrlich@setlabs.de} \\
}

\begin{document}
\maketitle
\begin{abstract}
Healthcare robotics requires robust multimodal perception and reasoning to ensure safety in dynamic clinical environments. Current Vision-Language Models (VLMs) demonstrate strong general-purpose capabilities but remain limited in temporal reasoning, uncertainty estimation, and structured outputs needed for robotic planning. We present a lightweight agentic multimodal framework for video-based scene understanding. Combining the Qwen2.5-VL-3B-Instruct model with a SmolAgent-based orchestration layer, it supports chain-of-thought reasoning, speech–vision fusion, and dynamic tool invocation. The framework generates structured scene graphs and leverages a hybrid retrieval module for interpretable and adaptive reasoning. Evaluations on the Video-MME benchmark and a custom clinical dataset show competitive accuracy and improved robustness compared to state-of-the-art VLMs, demonstrating its potential for applications in robot-assisted surgery, patient monitoring, and decision support.

\end{abstract}


\section{Introduction}
Robotics in healthcare has emerged as a critical domain where perception, reasoning, and safe decision-making intersect with high-stakes clinical applications. From robot-assisted surgery \cite{yang2017medical}, to autonomous patient monitoring \cite{davoudi2019intelligent}, and collaborative care robots \cite{fong2003survey}, the demand for systems that can robustly interpret complex multimodal environments continues to grow. A central requirement across these applications is scene understanding—the ability to identify objects, infer spatial and temporal relations, and generate structured representations that inform safe robotic actions \cite{zhou2024solving}.

\begin{figure}[h!]
    \centering
    \includegraphics[width=0.9\textwidth]{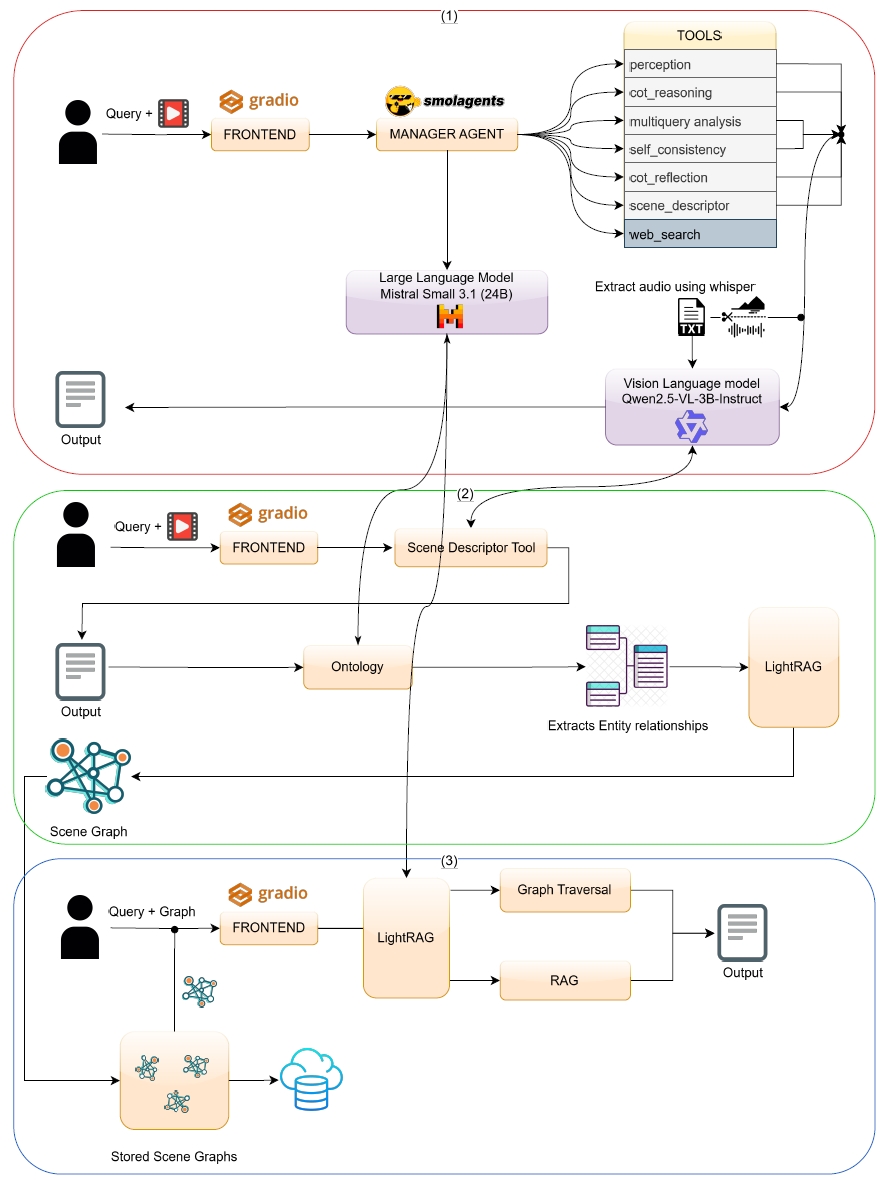}
    \caption{System architecture. (1) VisionQA
(highlighted in red), (2) SceneGen (highlighted in green), and (3) GraphQA (highlighted in blue). Each workflow represents a distinct method in the system's pipeline, with which a user can interact.}
    \label{fig:klinik_architecture}
\end{figure}

Recent advances in Vision-Language Models (VLMs) such as Llava \cite{liu2024llavanext} demonstrate strong multimodal reasoning capabilities. These systems have achieved remarkable performance in tasks such as visual question answering, image captioning, and document analysis. However, their application in robotics, especially in healthcare, faces critical challenges. First, most VLMs operate as monolithic end-to-end pipelines, limiting flexibility, explainability, and integration with robotic control loops \cite{chen2025standalone}. Second, temporal and spatial reasoning remains underdeveloped, hindering the accurate interpretation of dynamic surgical or clinical environments \cite{liu2025visual}. Third, the lack of structured outputs (e.g. scene graphs) complicates downstream integration into robotic planning frameworks \cite{liu2024llavanext,liu2025multi}. Finally, deployment in clinical workflows is often obstructed by the high computational cost of large proprietary models and unresolved concerns about data privacy \cite{alsaad2024multimodal}.

Within the healthcare domain, these shortcomings are especially problematic. For example, surgical robots require interpretable scene representations to coordinate tool trajectories with evolving anatomical contexts. Assistive robots in hospitals must combine perception with reasoning to safely navigate dynamic environments involving patients and caregivers. Clinical decision-support robots further demand uncertainty estimation and fallback strategies to avoid unsafe actions under ambiguity.

To address these challenges, we present a leightweight modular agentic multimodal framework designed for video-based scene understanding in clinical and robotic contexts. By integrating the Qwen2.5-VL-3B-Instruct model with a SmolAgent-based orchestration layer, our framwork combines chain-of-thought reasoning, speech-vision fusion, and structured scene graph generation. In addition, a hybrid retrieval mechanism (LightRAG \cite{guo2024lightrag}) enables both efficient and interpretable knowledge integration. Together, these components aim to bridge the gap between raw perception and symbolic planning, enabling healthcare robots to act more safely, transparently, and adaptively in high-stakes environments.

\section{RELATED WORK}
Vision-Language Models (VLMs) such as CLIP \cite{radford2021learning}, Flamingo \cite{alayrac2022flamingo}, and GPT-4V \cite{achiam2023gpt} demonstrate strong multimodal reasoning, but remain limited in temporal and spatial reasoning \cite{liu2025multi}. More recent approaches like GPT4Scene \cite{qi2025gpt4scene} or PRIME AI \cite{georgenthum2025enhancing} extend VLMs toward video and surgical contexts. However, these remain largely monolithic architectures with limited adaptability to robotic pipelines.

Agent-based frameworks such as ReAct \cite{yao2023react} and SmolAgents \cite{smolagents} emphasize modular reasoning, planning, and dynamic tool invocation. In video reasoning, approaches like Agent-of-Thoughts \cite{shi2025enhancing} and ViQAgent \cite{montes2025viqagent} demonstrate that incorporating external verification and multi-step reasoning improves robustness and interpretability. While promising, these systems are rarely applied to high-stakes domains such as healthcare robotics, where uncertainty handling and safety mechanisms are paramount \cite{alsaad2024multimodal}.

Scene graphs provide a structured and interpretable representation of objects and their relations, making them valuable for bridging perception and robotic control \cite{krishna2017visual}. Recent work in multimodal scene graph generation has shown strong potential in human–robot collaboration tasks, allowing robots to ground instructions and adapt to dynamic environments \cite{liu2024llavanext}. Despite these advances, scene graphs remain underutilized in medical robotics, where they could support interpretable surgical guidance, workflow modeling, and safe planning.

Robotics in healthcare spans from robot-assisted surgery \cite{yang2023improving} to collaborative assistive systems \cite{fong2003survey}. Evaluation benchmarks tailored for clinical domains are emerging, such as MedFrameQA \cite{yu2025medframeqa}, which emphasizes temporal reasoning over medical imagery. These efforts highlight the need for domain-specific testing to ensure safety and reliability. Yet, most VLMs and multimodal agents are not systematically validated in clinical robotics contexts, leaving a gap for frameworks designed with healthcare applicability in mind.
\newline
\par
In summary, prior research has advanced multimodal perception, agent-based reasoning, and structured scene representations. However, existing approaches remain either monolithic, poorly adapted to dynamic robotic pipelines, or insufficiently validated in healthcare contexts.

\section{Methods}

We propose a lightweight agentic multimodal framework for video-based scene understanding in healthcare robotics. The framework bridges perception and planning by combining a vision–language backbone, agentic orchestration, structured scene graph generation, and hybrid retrieval. Figure~\ref{fig:klinik_architecture} illustrates the system. Our framework was designed with two key considerations in mind: (i) supporting clinicians and robotic systems in dynamic, safety-critical environments, and (ii) ensuring interpretability and traceability of decisions, both of which are essential for deployment in healthcare contexts.

\subsection{System Overview}
Our framework is organized around three complementary workflows: \textit{VisionQA}, \textit{SceneGen}, and \textit{GraphQA}. VisionQA provides direct natural language interaction with video data, enabling clinicians to pose questions about ongoing procedures. SceneGen automatically transforms raw multimodal outputs into structured scene graphs, which serve as interpretable world models. GraphQA extends this capability by enabling reasoning directly over the structured representation, supporting symbolic queries such as tool–object interactions or temporal event ordering. 

These workflows are orchestrated by a lightweight agent layer that coordinates tasks, manages external knowledge access, and ensures consistent outputs. The modularity of this design directly addresses the high uncertainty and variability of clinical environments, where new tools, procedures, or workflows may appear unexpectedly. Moreover, the explicit orchestration enhances adaptability, enabling our framework to serve as both a perceptual assistant and a planning interface for robotic systems.

\subsection{Vision-Language Backbone}
At the core of our framework lies the Qwen2.5-VL-3B-Instruct model \cite{Qwen2.5-VL}, a transformer-based vision–language system optimized for multimodal instruction following. The model was selected after a systematic evaluation of eight contemporary VLMs (e.g., BLIP, Molmo, Moondream2, LLaVA-NeXT). Selection criteria included accuracy, multimodal capability, computational efficiency, and maintainability. Qwen2.5-VL-3B demonstrated the best trade-off between performance and resource demands, outperforming lighter models while avoiding the prohibitive VRAM requirements of larger ones (7B+).

The backbone employs an adaptive frame sampling strategy, selecting key frames based on motion cues and scene changes, thereby avoiding redundant computation while capturing clinically relevant events (e.g., introduction of a surgical tool). Each frame is paired with a task-specific prompt and processed to yield grounded captions, object descriptions, and candidate relations. To preserve temporal context, our framework employs a temporal memory buffer which aggregates output from consecutive frames. This mechanism enables reasoning about evolving surgical events (e.g., tool usage before vs. after incision) without requiring large-scale video pretraining, thus reducing computational overhead.

\subsection{Multimodal Processing}
To support audiovisual comprehension, our framework further integrates speech-to-text (STT) via Whisper \cite{radford2023robust}, combined with MoviePy \cite{zulko2020moviepy} for audio extraction and SpeechRecognition \cite{zhang2023speechrecognition} for interface standardization. This allows seamless integration of verbal instructions, procedural commentary, and background conversations into the multimodal reasoning pipeline. Whisper’s robustness in noisy environments ensures that domain-specific vocabulary is reliably captured, while GPU acceleration enables near real-time transcription.

The processing pipeline applies several optimizations to maintain efficiency in clinical contexts:
(i) Dynamic resolution handling (feature of the Qwen model) adapts input sizes to available GPU memory. (ii) FlashAttention-2 reduces inference latency and VRAM consumption. (iii) Caching mechanisms prevent redundant model loading during prolonged sessions. (iv) Accelerate-based multi-GPU execution supports scalability.

Together, these steps allow our system to process multi-minute video sequences on mid-range GPUs while maintaining performance.

\subsection{Agentic Orchestration}
While VLMs demonstrate strong pattern recognition, they often fail on multi-step reasoning or knowledge-intensive tasks. Our framework addresses this through a SmolAgent-based orchestration layer \cite{smolagents}, which follows the ReAct paradigm \cite{yao2023react}. Instead of producing a single response, the agent alternates between explicit reasoning steps and tool invocations.

This design provides three key advantages: (i) Modular Task Decomposition: Breaks complex tasks into manageable subtasks for better adaptability and robustness, (ii) Dynamic Tool Activation: Activates only necessary tools based on task requirements, ensuring efficient resource use, and (iii) explicit uncertainty handling, where fallback strategies are triggered if confidence in model output is low.  

By externalizing reasoning, the orchestration layer improves transparency and supports auditable interaction protocols, which are essential for robotic systems intended to assist in high-stakes environments such as operating rooms.

\subsection{Scene Graph Generation}
Structured representations are generated through the SceneGen module. Detected entities from VisionQA outputs are mapped to canonical object categories (e.g., “scalpel,” “forceps,” “tissue region”) using the Mistral small 3.1 LLM. Relations are inferred using a hybrid approach: a lightweight attention-based classifier for spatial relations (e.g., “above,” “next to”) and rule-based templates for temporal sequences (e.g., “before,” “after”). 

\begin{figure}[h!]
    \centering
    \includegraphics[width=0.6\columnwidth]{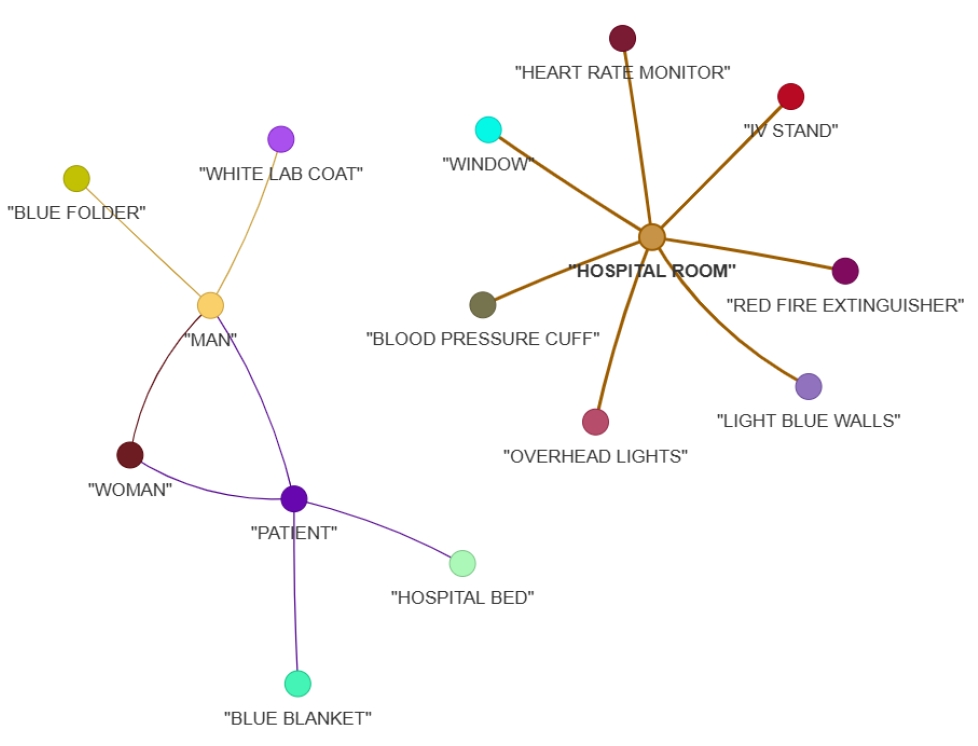}
    \caption{LightRAG based scene graph generation for structured and actionable knowledge representation.}
    \label{fig:graph}
\end{figure}

The resulting scene graph represents objects as nodes and relations as labeled edges, providing an interpretable abstraction of the environment (see example in Figure \ref{fig:graph}). These graphs not only facilitate explainable outputs for clinicians but also serve as state representations for robotic planners, allowing robots to align actions with the semantic structure of the surgical field. By bridging perception with symbolic representations, our framework addresses one of the core challenges in healthcare robotics: transforming raw video input into actionable, structured knowledge.

\subsection{GraphQA: Structured Reasoning}
Beyond perception, our system enables structured reasoning through the GraphQA module. For example, the query ``Which instrument most recently contacted the tissue?'' is translated into a search for edges linking tools and tissue nodes with the latest temporal label. This process grounds abstract questions in explicit graph operations, ensuring answers are both interpretable and verifiable.

This structured reasoning has two benefits: (i) robustness against hallucinations common in large language models, and (ii) traceability, since results can be directly linked to graph elements. This makes GraphQA particularly valuable in clinical robotics, where decision support must be auditable and defensible to meet safety and regulatory requirements.

\subsection{Hybrid Retrieval with LightRAG}
Our system further integrates LightRAG \cite{guo2024lightrag}, a retrieval framework that combines dense vector similarity with graph-based indexing. For instance, when asked about the recommended handling of a specific tool, the system retrieves relevant surgical guidelines and integrates them into the reasoning process. By combining high-level and low-level retrieval, the framework avoids purely opaque embeddings and ensures that retrieved content can be traced to authoritative references. This property is essential for healthcare robotics, where accountability and safety standards are strict.

\subsection{Interaction Modalities}
Finally, it supports two primary interaction modes: \textit{VisionQA}, where clinicians or operators interact via natural language and receive immediate answers to visual queries; and \textit{SceneGen}, where scene graphs or symbolic representations are exported for integration with robotic task planners or monitoring systems. These modes ensure that the framework can support both human-in-the-loop decision-making and automated robotic functions, bridging the gap between perception and action in healthcare robotics.

\section{EXPERIMENTS}

The evaluation of our system constitutes a critical component of this work, systematically assessing the system’s capabilities in scene understanding and reasoning, with a particular focus on the medical domain. The evaluation serves three main objectives: first, to validate technical performance against established multimodal benchmarks; second, to assess clinical relevance in realistic healthcare scenarios; and third, to identify specific strengths and limitations that inform both near-term applications and avenues for future improvement. Importantly, the assessment goes beyond conventional accuracy metrics and provides a multidimensional analysis of the framework’s suitability for real-world deployment.

To address both general and domain-specific aspects, we adopted a two-phase evaluation strategy. 

\subsection{Benchmark Validation using Video-MME Dataset}
In the first phase, our framework was benchmarked on the Video-MME dataset \cite{fu2025video}, a standardized suite that enables reproducible comparison across six visual domains and twelve task types. This phase establishes the system's performance relative to state-of-the-art multimodal systems, but its general-purpose nature necessitates complementary domain validation. We compared our framework to established vision–language models including GPT-4o, ByteVideoLLM, and VideoLlaVA, with all systems evaluated in a zero-shot setting. It relied on the Qwen2.5-VL-3B-Instruct model as backbone, extended through SmolAgent orchestration. Performance was measured using task-specific answer accuracy, complemented by inference speed as an indicator of computational efficiency.

To contextualize the system’s performance, we establish a baseline using the standard Qwen2.5-VL-3B-Instruct model in its default configuration. This baseline excludes all architectural enhancements and task-specific optimizations introduced, providing a reference point for quantifying the benefits of our agentic orchestration, scene graph generation, and retrieval modules. Both models were evaluated on the complete Video-MME test set (400 questions) under identical conditions to ensure comparability.

\subsection{Clinical Validation on Healthcare Scenarios}
In the second phase, we performed a clinical evaluation using a custom dataset of 20 medical videos and 80 annotated question–answer pairs. The dataset construction began with 40 candidate videos obtained from medical stock footage platforms \cite{mixkit2023hospital,pexels2023hospital,kirievskiy2022hospital}. From these, 20 videos were selected after multi-criteria filtering to ensure diversity, clinical relevance, and visual quality. The final collection covered a range of scenarios, including emergency care, diagnostic procedures, and common hospital workflows, with each video restricted to a short duration of up to one minute. 

To generate the evaluation set, four independent annotators—without involvement in system development—created 80 diverse question–answer pairs. These questions covered a spectrum of task categories, such as instrument identification, procedure classification, spatial and temporal reasoning, and attribute recognition. Both objective (e.g., counting tasks) and open-ended questions were included, ensuring evaluation across multiple reasoning modalities. Importantly, free-form questions prevented the model from exploiting syntactic biases common in templated datasets, instead requiring genuine contextual reasoning. Prior studies have shown \cite{krishna2017visual} that human-authored datasets reflect real-world reasoning processes more accurately than automatically generated ones, which motivated our reliance on manual annotation.

To evaluate model outputs, we combined human-provided ground truth with an LLM-based judgment framework. Specifically, DeepSeek-V3 \cite{guo2025deepseek} served as a secondary evaluator, comparing our system’s responses against reference answers. For multiple-choice questions, correctness was judged by direct match, while for open-ended queries, semantic equivalence was assessed, allowing flexibility in phrasing while maintaining factual correctness. This approach not only improved scalability and consistency but also enabled granular error analysis through model-generated justifications, following recent best practices for LLM-assisted evaluation \cite{zheng2023judging}.

\section{RESULTS}

\subsection{Benchmark Validation using Video-MME Dataset}
On the Video-MME benchmark \cite{fu2025video}, our framework achieved an overall accuracy of 70.5\% across 400 questions spanning 12 task categories (see Table \ref{tab:task_performance}). Performance varied by task type: the system excelled in Optical Character Recognition - OCR (95.7\%), information synopsis (95.2\%), and attribute perception (77.8\%), while more challenging tasks such as counting (41.1\%) and temporal reasoning (50.0\%) highlighted areas for improvement. Importantly, heavily weighted categories such as object recognition (65.4\%) and action recognition (60.0\%) showed consistent gains compared to the baseline model, contributing significantly to the overall performance.

\begin{table}[h!]
    \centering
    \caption{Performance by task type on the Video-MME benchmark.}
    \label{tab:task_performance}
    \begin{tabular}{lcc}
        \hline
        \textbf{Task Category} & \textbf{Correct/Total} & \textbf{Accuracy (\%)} \\
        \hline
        Action Reasoning      & 11/15  & 73.3 \\
        Action Recognition    & 33/55  & 60.0 \\
        Attribute Perception  & 42/54  & 77.8 \\
        Counting Problem      & 21/51  & 41.1 \\
        Information Synopsis  & 40/42  & 95.2 \\
        OCR Problems          & 22/23  & 95.7 \\
        Object Reasoning      & 30/38  & 78.9 \\
        Object Recognition    & 51/78  & 65.4 \\
        Spatial Perception    & 8/11   & 72.7 \\
        Spatial Reasoning     & 11/16  & 68.8 \\
        Temporal Perception   & 9/13   & 69.2 \\
        Temporal Reasoning    & 2/4    & 50.0 \\
        \hline
        \textbf{Average}      & 282/400 & \textbf{70.5} \\
        \hline
    \end{tabular}
\end{table}

\begin{figure}[!t]
    \centering
    \includegraphics[width=0.6\columnwidth]{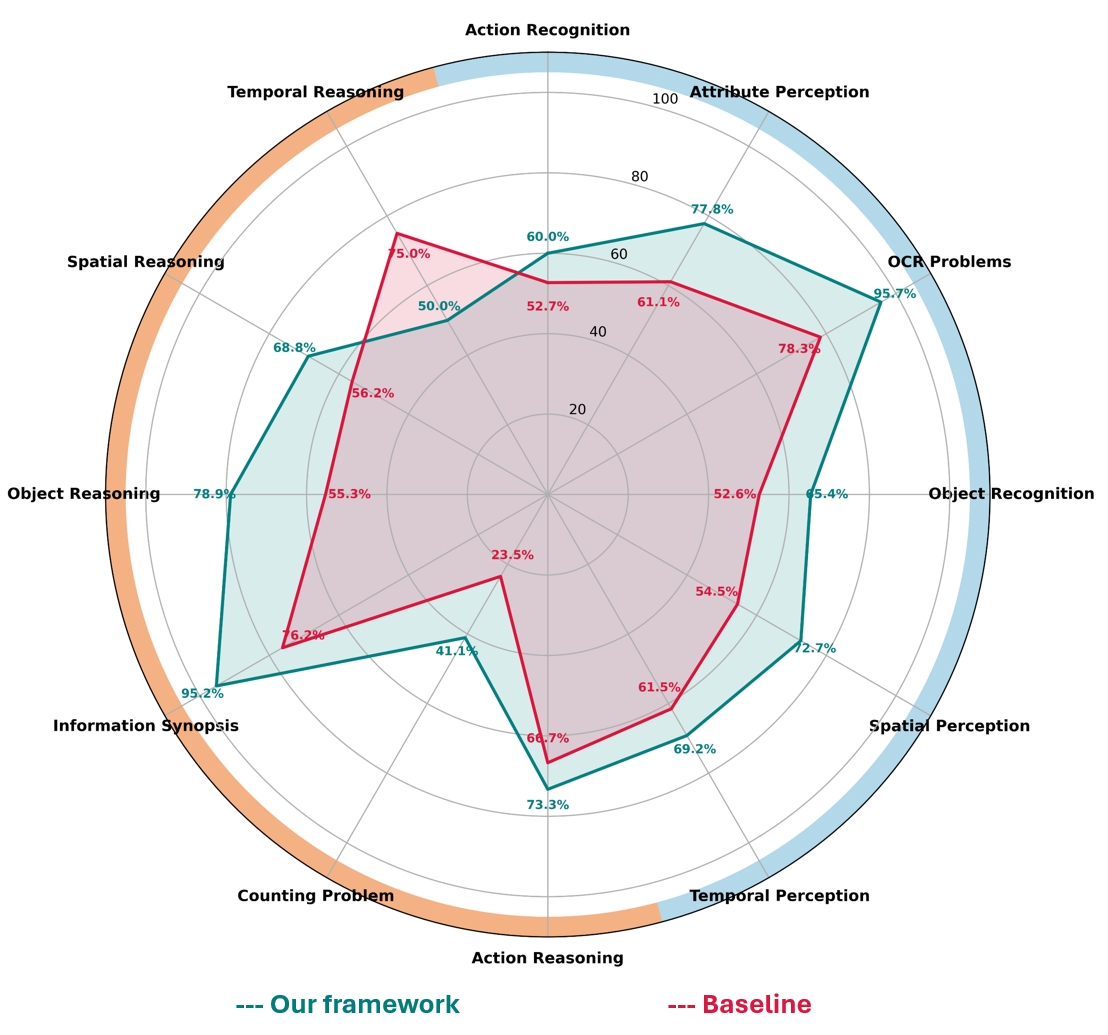}
    \caption{Performance by task type on the Video-MME benchmark comparing baseline (Qwen2.5-VL-3B-Instruct).}
    \label{fig:performance}
\end{figure}

When compared against the baseline Qwen2.5-VL-3B-Instruct, our framework achieves a 15\% absolute accuracy improvement, raising performance from 55.0\% to 70.0\% weighted accuracy. Notably, the improvements are concentrated in high-impact categories such as Object Reasoning (+23.6\%), Information Synopsis (+19.0\%), and Attribute Perception (+16.7\%), which collectively contribute substantially to the overall weighted accuracy (see Figure \ref{fig:performance}). Importantly, these gains were achieved without increasing model size, as both our system and the baseline use the same 3B-parameter backbone. This demonstrates that the observed improvements stem from the agentic orchestration and structured reasoning components rather than brute-force scaling.

Our framework also reduced performance variance across tasks (SD = 15.2 vs. 15.8 for the baseline) and improved lower-quartile accuracy, indicating stronger consistency in mid-tier task types. The skewness of results further showed that most tasks clustered at higher performance levels, with only a few outliers lowering the overall average.

Domain-level analysis confirmed that our system generalizes well across heterogeneous settings. Highest scores were obtained in Travel (89.5\%), Biology \& Medicine (88.9\%), and Technology (88.9\%), while domains such as Humanity \& History (40.0\%) and Magic Show (42.9\%) presented greater challenges.

When compared to state-of-the-art models on the Video-MME leaderboard (see Table \ref{tab:videomme_comparison}), our system (70.5\%) outperformed similarly sized open-weight systems such as Video-XL (64.0\%) and VideoChat2-Mistral (48.3\%), while approaching the performance of larger models such as ByteVideoLLM (74.4\%, 14B parameters). Although proprietary systems like InternVL2.5 (82.8\%) and Gemini 1.5 Pro (81.7\%) still lead, it demonstrates competitive efficiency, delivering nearly 85\% of ByteVideoLLM’s performance with only 21\% of the parameters.

These findings establish our framework as a parameter-efficient yet effective framework for multimodal scene understanding in healthcare-relevant settings, while also highlighting priority areas for improvement in object recognition, counting tasks, and temporal reasoning.

\begin{table}[!t]
    \centering
    \caption{Performance Comparison on Video-MME Benchmark (short-answer tasks)}
    \label{tab:videomme_comparison}
    \begin{tabular}{lcc}
    \hline
        \textbf{Model} & \textbf{LLM Parameters (B)} & \textbf{Score} \\
        \hline
        InternVL2.5          & 72  & 82.8 \\
        Gemini 1.5 Pro       & --  & 81.7 \\
        GPT-4o               & --  & 80.0 \\
        ByteVideoLLM         & 14  & 74.4 \\
        \textbf{Our framework} & \textbf{3}   & \textbf{70.5} \\
        Video-XL             & 7   & 64.0 \\
        Video-CCAM           & 14  & 62.2 \\
        InternVL-Chat-V1.5   & 20  & 60.2 \\
        VideoChat2-Mistral   & 7   & 48.3 \\
        Video-LLaVA          & 7   & 45.3 \\
        \hline
    \end{tabular}
\end{table}

\subsection{Clinical Validation on Healthcare Scenarios}

The evaluation of our framework on the curated medical dataset revealed robust performance in clinically relevant video-based reasoning tasks (see Table \ref{tab:custom_performance}). Across 80 annotated queries spanning eight question types, the system achieved an overall accuracy of 78.8\% (63/80 correct).

Performance was particularly strong in Temporal Reasoning (100\%), Information Synopsis (88.9\%), and both Action and Object Recognition (88.2\%), highlighting the system’s capacity to interpret medical workflows, recognize specialized equipment, and differentiate between clinical roles. These strengths underline the system’s ability to generalize effectively in realistic healthcare scenarios.

Challenging areas emerged in Counting Problems (63.6\%) and OCR tasks (50.0\%), where errors often stemmed from crowded clinical environments, partial occlusions, or unreliable text extraction from video frames. Attribute Perception (68.4\%) and Spatial Reasoning (66.7\%) showed moderate performance, suggesting limitations in fine-grained discrimination of visual details.

Taken together, the custom dataset evaluation demonstrates that our system provides reliable medical scene understanding while revealing actionable limitations. These results complement the general-purpose Video-MME benchmark, offering domain-specific insights essential for guiding future system improvements in healthcare applications.

\begin{table}[t]
    \centering
    \caption{Performance by question type on the custom medical dataset.}
    \label{tab:custom_performance}
    \begin{tabular}{lcc}
        \hline
        \textbf{Question Type} & \textbf{Number} & \textbf{Accuracy (\%)} \\
        \hline
        Counting Problems     & 11 & 63.6 \\
        Action Recognition    & 17 & 88.2 \\
        Object Recognition    & 17 & 88.2 \\
        Attribute Perception  & 19 & 68.4 \\
        Spatial Reasoning     & 3  & 66.7 \\
        Information Synopsis  & 9  & 88.9 \\
        OCR Problems          & 2  & 50.0 \\
        Temporal Reasoning    & 2  & 100.0 \\
        \hline
    \end{tabular}
\end{table}

\section{DISCUSSION}

This work introduced a lightweight agentic multimodal framework for clinical scene understanding that combines vision–language reasoning, structured scene graph generation, and retrieval-augmented question answering. By validating the system both on the Video-MME benchmark and on a custom medical dataset, we demonstrated that our framwork can achieve competitive accuracy while producing interpretable outputs that are directly useful for both clinicians and robotic systems. These findings highlight the potential of parameter-efficient multimodal models to deliver clinically relevant insights without relying on prohibitively large architectures.  

At the same time, the evaluation also revealed areas where the current design can be further strengthened. The reliance on iterative agentic reasoning, while valuable for transparency, introduces latency that limits real-time applicability on long video sequences or constrained hardware. Similarly, the absence of dedicated medical training data means that rare procedures and specialized instruments are not always recognized with high reliability. The system has not yet been stress-tested under challenging conditions common in hospitals, such as occlusions, variable lighting, or background noise. These limitations should not be seen as obstacles but rather as directions for refinement.  

Several promising paths emerge from this work. Integrating explicit temporal reasoning into the scene graph could transform the system from a frame-level interpreter into a system capable of monitoring workflows and anticipating procedural steps. Exploring multi-agent orchestration offers a way to balance specialized reasoning with collaborative decision-making, aligning closely with the teamwork dynamics of real clinical practice \cite{wang2024videoagent}. Domain adaptation strategies, supported by carefully curated datasets and incremental learning from real-world interactions, can make the system more attuned to specific medical contexts while preserving privacy \cite{guo2023survey, mai2024efficient}. Equally important, advances in confidence estimation \cite{liu2025multi} and sensor fusion will allow the framework to communicate uncertainty more effectively—an essential quality in high-stakes environments. Finally, closing the loop with interactive clinician feedback promises not only to improve accuracy over time but also to embed the system naturally into daily healthcare workflows.  

Taken together, these insights outline a path toward multimodal AI systems that are interpretable, adaptable, and collaborative. By supporting both clinicians and robotic assistants, it moves closer to a vision of human–AI partnership in medicine where safety, transparency, and efficiency go hand in hand.

\section{CONCLUSION}

This work introduced a lightweight agentic multimodal framework for video-based scene understanding in healthcare robotics. By integrating a vision--language backbone, agentic orchestration, structured scene graph generation, and hybrid retrieval, the system bridges perception and reasoning in complex medical environments. Quantitative and qualitative evaluations demonstrate that the system achieves competitive benchmark performance while offering domain-specific interpretability, making it suitable for both research and clinical exploration. While still a prototype, our approach underscores the potential of agentic multimodal AI to support safe and transparent human–robot collaboration. In addition, our work illustrates how interpretable agentic multimodal AI can jointly support clinicians and robotic systems, contributing to the development of more trustworthy and collaborative decision support in healthcare.

\section*{ACKNOWLEDGMENT}

The authors acknowledge the financial support within
the COMET K2 Competence Centers for Excellent Technologies
from the Austrian Federal Ministry for Climate
Action (BMK), the Austrian Federal Ministry for Labour and
Economy (BMAW), the Province of Styria (Dept. 12) and
the Styrian Business Promotion Agency (SFG). The Austrian
Research Promotion Agency (FFG) has been authorized for
the program management. Disclosure of use of AI tools: OpenAI GPT-5 was used for text refinements and generation of tables.

\bibliographystyle{unsrt}  


\begin{thebibliography}{10}

\bibitem{yang2017medical}
G.-Z. Yang, J.~Cambias, K.~Cleary, E.~Daimler, J.~Drake, P.~E. Dupont, N.~Hata, P.~Kazanzides, S.~Martel, R.~V. Patel, {\em et~al.}, ``Medical robotics—regulatory, ethical, and legal considerations for increasing levels of autonomy,'' 2017.

\bibitem{davoudi2019intelligent}
A.~Davoudi, K.~R. Malhotra, B.~Shickel, S.~Siegel, S.~Williams, M.~Ruppert, E.~Bihorac, T.~Ozrazgat-Baslanti, P.~J. Tighe, A.~Bihorac, {\em et~al.}, ``Intelligent icu for autonomous patient monitoring using pervasive sensing and deep learning,'' {\em Scientific reports}, vol.~9, no.~1, p.~8020, 2019.

\bibitem{fong2003survey}
T.~Fong, I.~Nourbakhsh, and K.~Dautenhahn, ``A survey of socially interactive robots,'' {\em Robotics and autonomous systems}, vol.~42, no.~3-4, pp.~143--166, 2003.

\bibitem{zhou2024solving}
S.~Zhou, {\em Solving Real-World Tasks with AI Agents}.
\newblock PhD thesis, Carnegie Mellon University, 2024.

\bibitem{liu2024llavanext}
H.~Liu, C.~Li, Y.~Li, B.~Li, Y.~Zhang, S.~Shen, and Y.~J. Lee, ``Llavanext: Improved reasoning, ocr, and world knowledge,'' 2024.

\bibitem{chen2025standalone}
J.~Chen, J.~Ye, and G.~Wang, ``From standalone llms to integrated intelligence: A survey of compound al systems,'' {\em arXiv preprint arXiv:2506.04565}, 2025.

\bibitem{liu2025visual}
C.~Liu, Y.~Jin, Z.~Guan, T.~Li, Y.~Qin, B.~Qian, Z.~Jiang, Y.~Wu, X.~Wang, Y.~F. Zheng, {\em et~al.}, ``Visual--language foundation models in medicine,'' {\em The Visual Computer}, vol.~41, no.~4, pp.~2953--2972, 2025.

\bibitem{liu2025multi}
A.~Liu, W.~Jiang, S.~Huang, and Z.~Feng, ``Multi-modal integrated sensing and communication in internet of things with large language models,'' {\em IEEE Internet of Things Magazine}, 2025.

\bibitem{alsaad2024multimodal}
R.~AlSaad, A.~Abd-Alrazaq, S.~Boughorbel, A.~Ahmed, M.-A. Renault, R.~Damseh, and J.~Sheikh, ``Multimodal large language models in health care: applications, challenges, and future outlook,'' {\em Journal of medical Internet research}, vol.~26, p.~e59505, 2024.

\bibitem{guo2024lightrag}
Z.~Guo, L.~Xia, Y.~Yu, T.~Ao, and C.~Huang, ``Lightrag: Simple and fast retrieval-augmented generation,'' {\em arXiv preprint arXiv:2410.05779}, 2024.

\bibitem{radford2021learning}
A.~Radford, J.~W. Kim, C.~Hallacy, A.~Ramesh, G.~Goh, S.~Agarwal, G.~Sastry, A.~Askell, P.~Mishkin, J.~Clark, {\em et~al.}, ``Learning transferable visual models from natural language supervision,'' in {\em International conference on machine learning}, pp.~8748--8763, PmLR, 2021.

\bibitem{alayrac2022flamingo}
J.-B. Alayrac, J.~Donahue, P.~Luc, A.~Miech, I.~Barr, Y.~Hasson, K.~Lenc, A.~Mensch, K.~Millican, M.~Reynolds, {\em et~al.}, ``Flamingo: a visual language model for few-shot learning,'' {\em Advances in neural information processing systems}, vol.~35, pp.~23716--23736, 2022.

\bibitem{achiam2023gpt}
J.~Achiam, S.~Adler, S.~Agarwal, L.~Ahmad, I.~Akkaya, F.~L. Aleman, D.~Almeida, J.~Altenschmidt, S.~Altman, S.~Anadkat, {\em et~al.}, ``Gpt-4 technical report,'' {\em arXiv preprint arXiv:2303.08774}, 2023.

\bibitem{qi2025gpt4scene}
Z.~Qi, Z.~Zhang, Y.~Fang, J.~Wang, and H.~Zhao, ``Gpt4scene: Understand 3d scenes from videos with vision-language models,'' {\em arXiv preprint arXiv:2501.01428}, 2025.

\bibitem{georgenthum2025enhancing}
H.~Georgenthum, C.~Cosentino, F.~Marozzo, and P.~Li{\`o}, ``Enhancing surgical documentation through multimodal visual-temporal transformers and generative ai,'' {\em arXiv preprint arXiv:2504.19918}, 2025.

\bibitem{yao2023react}
S.~Yao, J.~Zhao, D.~Yu, N.~Du, I.~Shafran, K.~Narasimhan, and Y.~Cao, ``React: Synergizing reasoning and acting in language models,'' in {\em International Conference on Learning Representations (ICLR)}, 2023.

\bibitem{smolagents}
A.~Roucher, A.~V. del Moral, T.~Wolf, L.~von Werra, and E.~Kaunismäki, ```smolagents`: a smol library to build great agentic systems..'' \url{https://github.com/huggingface/smolagents}, 2025.

\bibitem{shi2025enhancing}
Y.~Shi, S.~Di, Q.~Chen, and W.~Xie, ``Enhancing video-llm reasoning via agent-of-thoughts distillation,'' in {\em Proceedings of the Computer Vision and Pattern Recognition Conference}, pp.~8523--8533, 2025.

\bibitem{montes2025viqagent}
T.~Montes and F.~Lozano, ``Viqagent: Zero-shot video question answering via agent with open-vocabulary grounding validation,'' {\em arXiv preprint arXiv:2505.15928}, 2025.

\bibitem{krishna2017visual}
R.~Krishna, Y.~Zhu, O.~Groth, J.~Johnson, K.~Hata, J.~Kravitz, S.~Chen, Y.~Kalantidis, L.-J. Li, D.~A. Shamma, {\em et~al.}, ``Visual genome: Connecting language and vision using crowdsourced dense image annotations,'' {\em International journal of computer vision}, vol.~123, no.~1, pp.~32--73, 2017.

\bibitem{yang2023improving}
Y.~Yang, H.~Li, Y.~Wang, and Y.~Wang, ``Improving the reliability of large language models by leveraging uncertainty-aware in-context learning,'' {\em arXiv preprint arXiv:2310.04782}, 2023.

\bibitem{yu2025medframeqa}
S.~Yu, H.~Wang, J.~Wu, C.~Xie, and Y.~Zhou, ``Medframeqa: A multi-image medical vqa benchmark for clinical reasoning,'' {\em arXiv preprint arXiv:2505.16964}, 2025.

\bibitem{Qwen2.5-VL}
S.~Bai, K.~Chen, X.~Liu, J.~Wang, W.~Ge, S.~Song, K.~Dang, P.~Wang, S.~Wang, J.~Tang, H.~Zhong, Y.~Zhu, M.~Yang, Z.~Li, J.~Wan, P.~Wang, W.~Ding, Z.~Fu, Y.~Xu, J.~Ye, X.~Zhang, T.~Xie, Z.~Cheng, H.~Zhang, Z.~Yang, H.~Xu, and J.~Lin, ``Qwen2.5-vl technical report,'' {\em arXiv preprint arXiv:2502.13923}, 2025.

\bibitem{radford2023robust}
A.~Radford, J.~W. Kim, T.~Xu, G.~Brockman, C.~McLeavey, and I.~Sutskever, ``Robust speech recognition via large-scale weak supervision,'' in {\em International conference on machine learning}, pp.~28492--28518, PMLR, 2023.

\bibitem{zulko2020moviepy}
Zulko, ``Moviepy: Python video editing library.'' \url{https://pypi.org/project/moviepy/}, 2020.
\newblock Python Package Index, version 1.0.3.

\bibitem{zhang2023speechrecognition}
A.~Zhang, ``Speechrecognition: Library for performing speech recognition.'' \url{https://pypi.org/project/SpeechRecognition/}, 2023.
\newblock Python Package Index, version 3.10.0.

\bibitem{fu2025video}
C.~Fu, Y.~Dai, Y.~Luo, L.~Li, S.~Ren, R.~Zhang, Z.~Wang, C.~Zhou, Y.~Shen, M.~Zhang, {\em et~al.}, ``Video-mme: The first-ever comprehensive evaluation benchmark of multi-modal llms in video analysis,'' in {\em Proceedings of the Computer Vision and Pattern Recognition Conference}, pp.~24108--24118, 2025.

\bibitem{mixkit2023hospital}
Mixkit, ``Free hospital stock videos.'' \url{https://mixkit.co/free-stock-video/hospital/}, 2023.
\newblock Stock Video Collection. Licensed under Mixkit Free License.

\bibitem{pexels2023hospital}
Pexels, ``Hospital stock videos.'' \url{https://www.pexels.com/search/videos/hospital/}, 2023.
\newblock Stock Video Collection. Licensed under Pexels License.

\bibitem{kirievskiy2022hospital}
A.~Kirievskiy, ``Hospital video footage,'' 2022.
\newblock Stock Video. Used under Creative Commons License.

\bibitem{guo2025deepseek}
D.~Guo, D.~Yang, H.~Zhang, J.~Song, R.~Zhang, R.~Xu, Q.~Zhu, S.~Ma, P.~Wang, X.~Bi, {\em et~al.}, ``Deepseek-r1: Incentivizing reasoning capability in llms via reinforcement learning,'' {\em arXiv preprint arXiv:2501.12948}, 2025.

\bibitem{zheng2023judging}
L.~Zheng, W.-L. Chiang, Y.~Sheng, S.~Zhuang, Z.~Wu, Y.~Zhuang, Z.~Lin, Z.~Li, D.~Li, E.~Xing, {\em et~al.}, ``Judging llm-as-a-judge with mt-bench and chatbot arena,'' {\em Advances in neural information processing systems}, vol.~36, pp.~46595--46623, 2023.

\bibitem{wang2024videoagent}
X.~Wang, Y.~Zhang, O.~Zohar, and S.~Yeung-Levy, ``Videoagent: Long-form video understanding with large language model as agent,'' in {\em European Conference on Computer Vision}, pp.~58--76, Springer, 2024.

\bibitem{guo2023survey}
R.~Guo, J.~Wei, L.~Sun, B.~Yu, G.~Chang, D.~Liu, S.~Zhang, Z.~Yao, M.~Xu, and L.~Bu, ``A survey on image-text multimodal models,'' {\em arXiv preprint arXiv:2309.15857}, 2023.

\bibitem{mai2024efficient}
X.~Mai, Z.~Tao, J.~Lin, H.~Wang, Y.~Chang, Y.~Kang, Y.~Wang, and W.~Zhang, ``From efficient multimodal models to world models: A survey,'' {\em arXiv preprint arXiv:2407.00118}, 2024.

\end{thebibliography}

\end{document}